\documentclass[letterpaper, 10 pt, conference]{ieeeconf}  % Comment this line out if you need a4paper

\IEEEoverridecommandlockouts                              % This command is only needed if 
\usepackage{graphicx}
\usepackage{epsfig} % for postscript graphics files
\usepackage{mathptmx} % assumes new font selection scheme installed
\usepackage{times} % assumes new font selection scheme installed
\usepackage{amsmath} % assumes amsmath package installed
\usepackage{amssymb}  % assumes amsmath package installed
\usepackage{url}
\title{\LARGE \bf
Unified Understanding of Environment, Task, and Human \\ for Human-Robot Interaction in Real-World Environments
}

\author{Yuga Yano$^{1}$, Akinobu Mizutani$^{1}$, Yukiya Fukuda$^{1}$, Daiju Kanaoka$^{1}$, Tomohiro Ono$^{1}$, and Hakaru Tamukoh$^{1, 2}$
\thanks{*This paper is based on results obtained from project JPNP16007 commissioned by the New Energy and Industrial Technology Development Organization (NEDO). This work was supported by JSPS KAKENHI Grant Numbers 23H03468 and 23K18495. This work was supported JST ALCA-Next Grant Number JPMJAN23F3.}
\thanks{All authors are with Kyushu Institute of Technology, Fukuoka, Japan.
        {\tt\small\{yano.yuuga158, mizutani.akinobu515, yukiya.fukuda265, kanaoka.daiju327, ono.tomohiro342\}@mail.kyutech.jp} and {\tt\small tamukoh@brain.kyutech.ac.jp}
        }
\thanks{Hakaru Tamukoh is also affiliated with the Research Center for Neuromorphic AI Hardware, Fukuoka Japan.}
}

\begin{document}

\maketitle
\thispagestyle{empty}
\pagestyle{empty}

\begin{abstract}
To facilitate human--robot interaction (HRI) tasks in real-world scenarios, service robots must adapt to dynamic environments and understand the required tasks while effectively communicating with humans.
To accomplish HRI in practice, we propose a novel indoor dynamic map, task understanding system, and response generation system.
The indoor dynamic map optimizes robot behavior by managing an occupancy grid map and dynamic information, such as furniture and humans, in separate layers.
The task understanding system targets tasks that require multiple actions, such as serving ordered items.
Task representations that predefine the flow of necessary actions are applied to achieve highly accurate understanding.
The response generation system is executed in parallel with task understanding to facilitate smooth HRI by informing humans of the subsequent actions of the robot.
In this study, we focused on waiter duties in a restaurant setting as a representative application of HRI in a dynamic environment.
We developed an HRI system that could perform tasks such as serving food and cleaning up while communicating with customers.
In experiments conducted in a simulated restaurant environment, the proposed HRI system successfully communicated with customers and served ordered food with 90\% accuracy.
In a questionnaire administered after the experiment, the HRI system of the robot received 4.2 points out of 5.
These outcomes indicated the effectiveness of the proposed method and HRI system in executing waiter tasks in real-world environments.

\end{abstract}

\begin{keywords}
Service robot, Dynamic map, Large language model, Real-World robotics
\end{keywords}

\section{Introduction}
\begin{figure}[t]
    \centering
    \includegraphics[width=\linewidth]{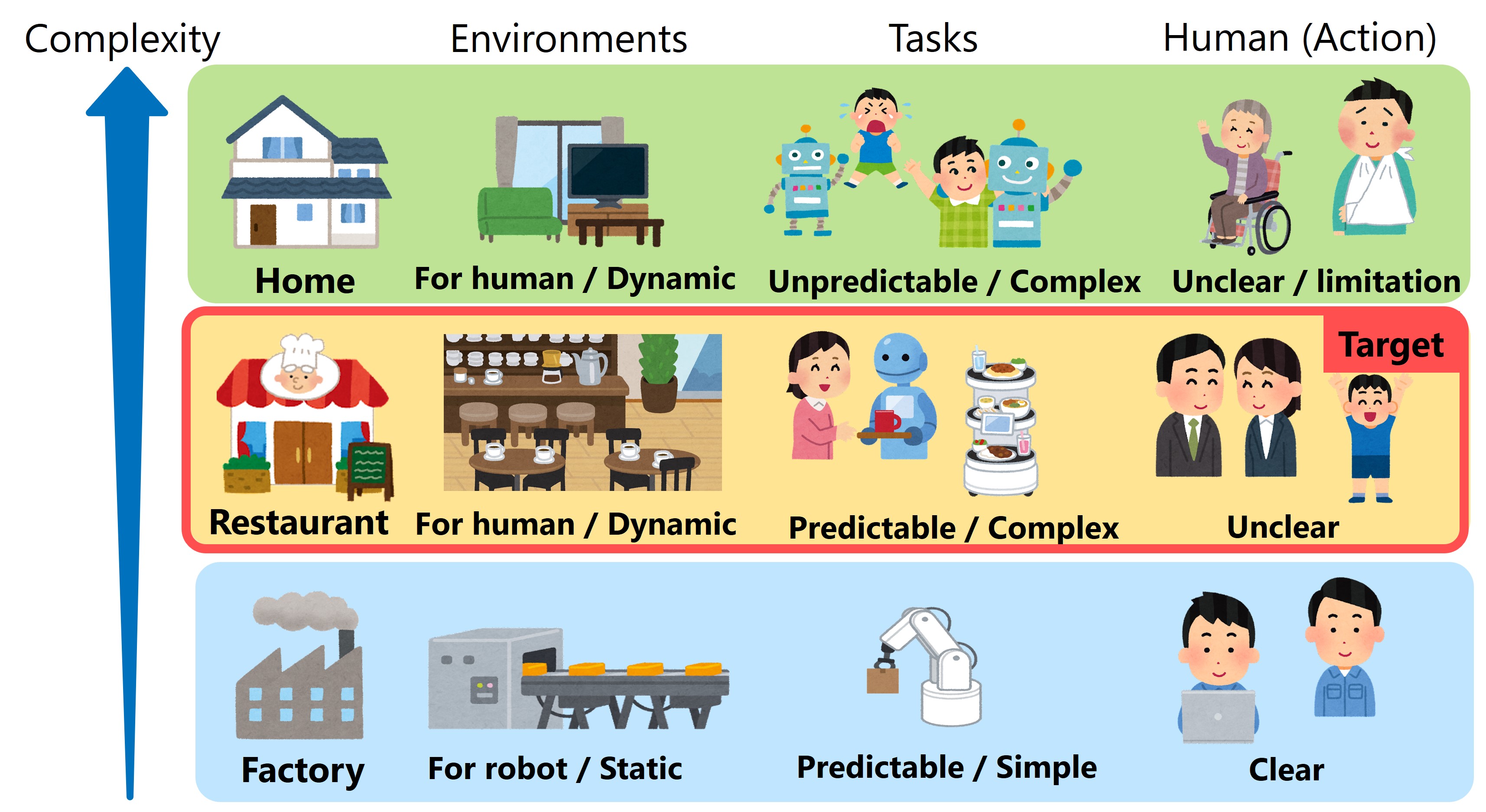}
    \caption{Complexity of real-world environment and our research target}
    \label{complexity}
\end{figure}

The integration of service robots that can understand and perform various tasks while communicating with people in restaurants and home environments is expected to improve the quality of human life~\cite{Waitersystem2023, Ono2022}.
Fig.~\ref{complexity} shows the differences in complexity and required tasks in each environment~\cite{realworld_robotics}.
Industrial robots typically operate in static environments designed for their functionality, often do not need to interact with humans, and only need to perform specific actions.
In contrast, service robots operate in diverse and dynamic environments in real-world conditions tailored to human activities.
To achieve human-robot interaction (HRI) tasks in real-world environments, robots must be able to adapt to dynamic environments and understand the required tasks while effectively interacting with humans.
Furthermore, the ability to understand humans and communicate with them by generating appropriate responses are crucial for achieving smooth HRI.

Simultaneous localization and mapping (SLAM), a widely used localization and mapping method, often uses an occupancy grid map.
However, occupancy grid maps are binary representations of the surrounding environment and cannot hold semantic information regarding objects.
Semantic SLAM~\cite{Rosinol19icra-incremental, Rosinol20icra-Kimera} can represent environments as occupancy grid maps, incorporating semantic information such as furniture.
Although multiple types of information can be represented on an occupancy grid map, recreation of the maps is challenging in real-world scenarios.
Additionally, it is difficult to assign multiple meanings to the same grid.
% To ensure simplified management, dynamic information, such as that of people and furniture, should be separated from the occupancy grid map.
To ensure simplified management, dynamic information, such as people and furniture, should be separated from the occupancy grid map.
Considering these issues, a dynamic map, which manages information based on an update rate in four layers, has been proposed for autonomous driving applications.
% However, there are few examples of its application in indoor environments \cite{mulit_layer}.
Based on this concept, we propose an indoor dynamic map containing necessary information to allow service robots to adapt to dynamic environments.
% In this study, we propose a dynamic map for the home environment to apply this dynamic map concept to the field of home service robots.
% In this study, we propose a dynamic map for indoor contained information to achieve more flexible HRI tasks.
% In this study, we propose an indoor dynamic map contained information to adapt to dynamic environments.

Furthermore, service robots need to perform various tasks through interaction with humans.
For example, the task of ``bringing orange juice'' involves multiple actions, such as determining a route, recognizing the juice, and planning to grasp it.
In certain cases, an operation may fail due to unforeseen events, or additional operations may be required.
Represent all these cases through rules is typically challenging.
% We cannot design all of these cases based on rules.
% In addition, there may also be cases where opening the refrigerator to get the orange juice occurs while the task is being executed or where the task cannot be completed because the orange juice is in a high place and cannot be grasped.
% We cannot design all of these cases based on rules.
% Therefore, it is necessary to design the most appropriate task for each situation.
Recently, SayCan~\cite{saycan2022arxiv} has been proposed for task understanding based on large language models (LLMs).
% SayCan \cite{saycan2022arxiv} defines 511 low-level behaviors (skills), such as “Find a coke” and “Grasp a sponge,” and predicts the order of skills to be executed to complete a command using PaLM.
% SayCan can execute simple commands such as "Pick up the coke can" in a known environment with a success rate of 100 \%, but for tasks that require the execution of multiple skills such as “Bring me a sponge” in an unknown environment, the accuracy drops to 73 \%.
Although this method can predict and execute simple commands with high accuracy, the accuracy and inference speed decrease as the tasks become more complex.
An affordance function that predicts the set of skills that a robot can perform based on input images can improve execution accuracy.
However, a large amount of robot demonstration data is necessary to learn the affordance function.
In this study, we propose a method to enhance the accuracy of understanding complex tasks without a large scale of robot demonstration data.

Fig.~\ref{complexity} shows the difference in complexity between restaurants and home environments.
Compared with a home environment, operating in a restaurant is simpler for service robots because of the fixed flow of operations to accomplish the required tasks.
Therefore, this study focuses on waiter tasks in a restaurant as a preliminary step to achieve autonomous behavior of robots in a complex home environment.
% Therefore, this study targets autonomous behavior in a restaurant as a preliminary step to achieve behavior in a complex home environment.
% We design a task that imitates a waiter's work in a restaurant and develop an HRI system that uses proposed dynamic map, understanding task and generating response system to realize the task.
Specifically, we design a task imitating a waiter's tasks in a restaurant and develop an HRI system that uses the proposed dynamic map, task-understanding, and response generation system to realize the task.
Social acceptance of service robots is also essential for their widespread use in real-world environments.
Therefore, we verify the social acceptance of the proposed HRI system through restaurant demonstrations for customers.

Our main contributions in this study are as follows:
% Our main contributions of this work can be summarized as follows:
% We propose an indoor dynamic map and implement multiple necessary actions to adapt to real-world environments.
% We introduce a task understanding system and response generation system utilizing LLMs to perform waiter tasks.
% We verify the social acceptance of the proposed HRI system based on the proposed methods.
\begin{itemize}
    \item We propose an indoor dynamic map and implement multiple necessary skills to adapt to real-world environments.
    \item We introduce a task understanding system and response generation system utilizing LLMs to perform waiter tasks.
    \item We verify the social acceptance of the proposed HRI system based on the proposed methods.
\end{itemize}

\section{Related works}

\subsection{Multiple layer environmental map}
\begin{figure}[t]
    \centering
    \includegraphics[width=0.85\linewidth]{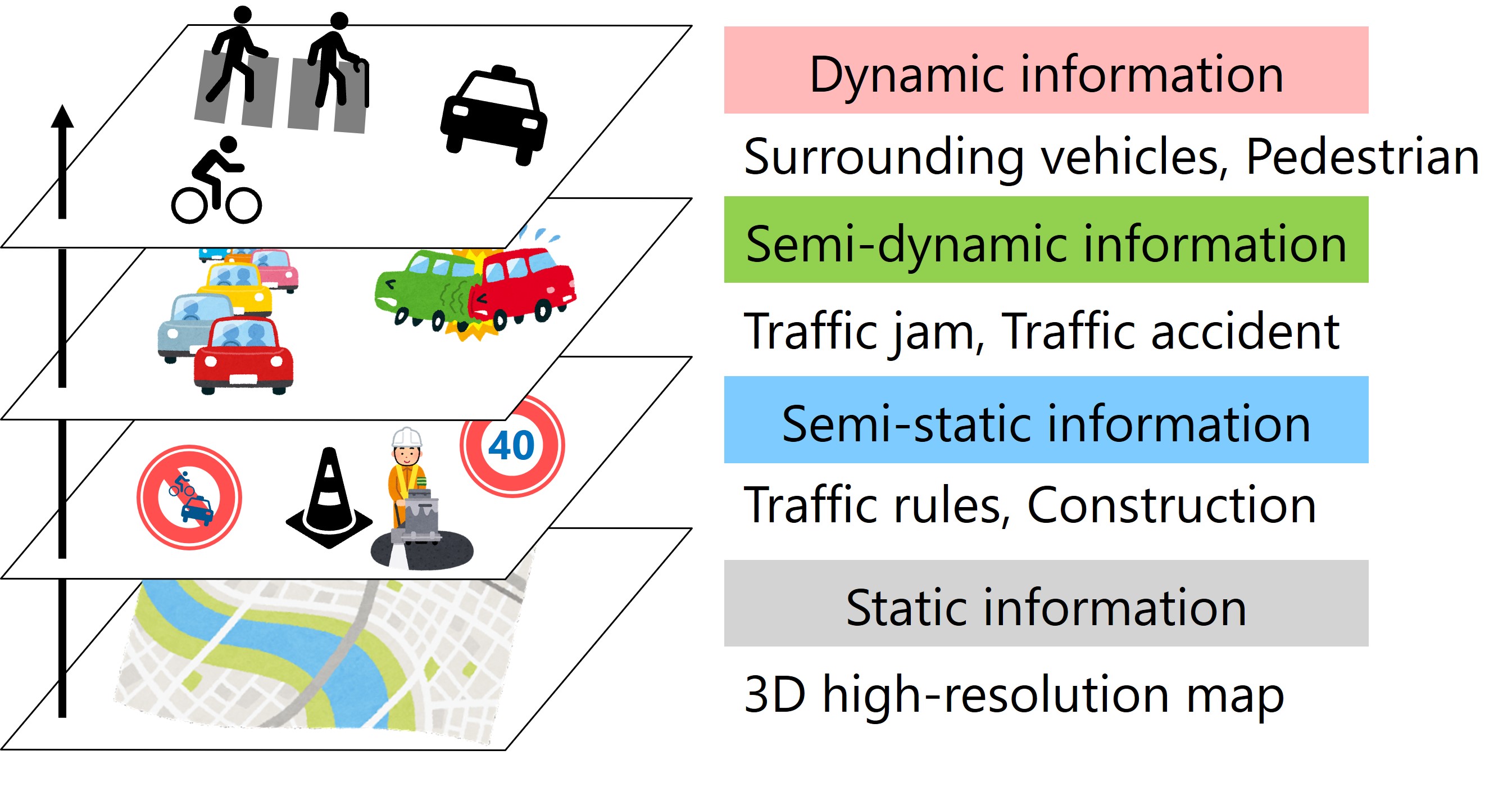}
    \caption{Overview of dynamic map in autonomous driving}
    \label{dynamic_map_orig}
\end{figure}

Fig.~\ref{dynamic_map_orig} shows a conceptual diagram of a dynamic map.
% The concept of a dynamic map, which consistently links static high-precision 3D map information with other dynamic data, such as traffic information, is used to advance automated driving.
Dynamic maps, which manage information in multiple layers, have proposed as an effective tool for advancing automated driving.
However, it is limited that research on indoor environmental maps based on multiple layers, such as dynamic maps.
Multi-layer environmental affordance map~\cite{mulit_layer} is an indoor map that manages information in static, obstacles, scene, and event layers.
This map manages information on furniture in the object layer and semantic information of rooms in the scene layer in addition to the grid map.
Additionally, this map have an event layer that detects user actions and events by integrating the results of human recognition.
This information helps the robot move safely and efficiently through the environment and generate more sophisticated navigation strategies.
However, this map cannot retain the shape of objects and thus cannot facilitate motion optimization.
In addition, this map uses YOLOv3 \cite{yolov3} for furniture recognition, which makes it challenging to adapt to unknown environments.

% \subsection{Understanding skills from natural language}
\subsection{Understanding tasks from natural language}
One of the significant challenges for the real-world application of service robots is understanding tasks from natural language.
For example, consider the command, ``I spilled some juice on the floor.''
Humans can understand the actions required to accomplish the desired task: 1. go to the kitchen, 2. find a sponge, 3. grab the sponge, 4. move near the human, 5. wipe the spill with the sponge.
However, it is difficult for robots to predict necessary actions from natural language instructions.
To solve this problem, Google proposed SayCan~\cite{saycan2022arxiv} for understanding tasks from instructions.
SayCan defines low-level actions (skills) that a robot can perform, such as moving, finding, and grasping, and decides which skill to execute based on the given instructions and the current situation.
In general, SayCan can accurately predict and execute instructions containing only one skill, such as ``How would you go to the table?'' with high accuracy.
However, when multiple skills are required to be executed, such as ``How would you bring me a Coke can,'' the prediction accuracy decreases.
An affordance function that predicts the set of skills that a robot can perform based on input images can improve task execution accuracy.
However, a large amount of robot demonstration data for reinforcement learning is necessary to acquire the affordance function.

\subsection{RoboCup@Home}
\begin{figure}[t]
    \centering
    \includegraphics[width=0.55\linewidth]{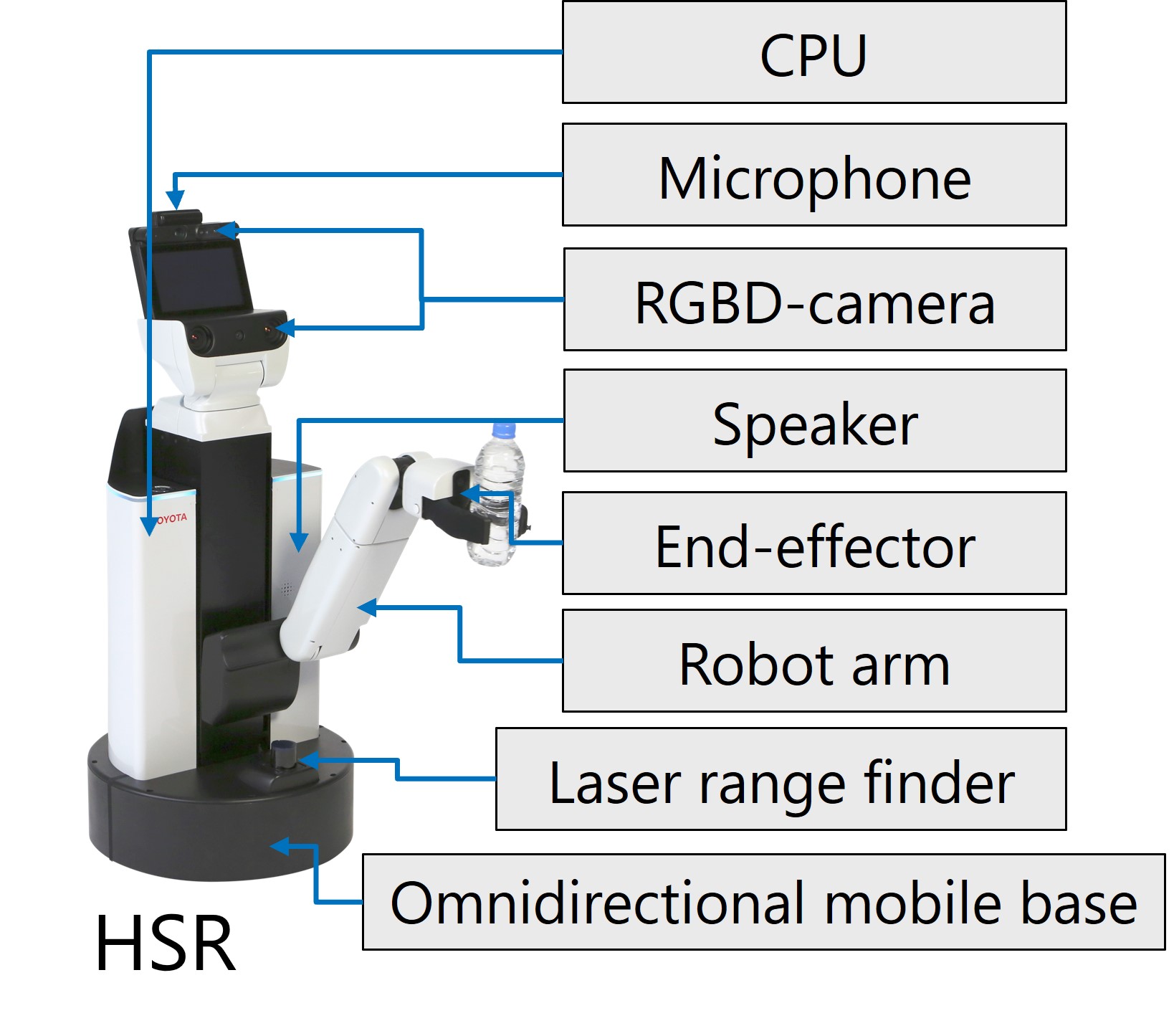}
    % \caption{Overview of HSR}
    \caption{Human Support Robot (HSR) and its main sensors and actuators}
    \label{hsr_overview}
\end{figure}

% \begin{figure}[t]
%     \centering
%     \includegraphics[width=0.55\linewidth]{figures/02_related_works/hsr_overview_yy3.jpg}
%     % \caption{Overview of HSR}
%     \caption{Human Support Robot (HSR) and its main sensors and actuators}
%     \label{hsr_overview}
% \end{figure}
RoboCup@Home~\cite{robocup} conducts a benchmark test named restaurant, which can evaluate the performance of waiter tasks using Human Support Robot (HSR)~\cite{yamamoto2019hsr} as shown in Fig.~\ref{hsr_overview}.
In this test, the robot autonomously finds customers, takes orders, and serves items in a restaurant.
The restaurant test is a challenging test with no team to complete because it must operate in an unknown environment where the environment and objects used for the test are unknown.
In order to solve the restaurant test, the customers' orders and dynamically changing environments must be effectively understood. 
The existing environmental map~\cite{mulit_layer} does not represent dynamic information in separate layers. Thus, its application to the restaurant test is challenging owing to the necessity of flexible adaptation to dynamic environments.
In addition, the map cannot be used to optimize robot behavior, such as object grasping, because it does not represent furniture shapes.
In terms of task understanding, SayCan cannot complete a task with sufficient accuracy because the restaurant test requires sequential execution of multiple actions, including movement, recognition, and grasping.
Learning the affordance function to improve the prediction accuracy of SayCan requires the acquisition of large-scale robot demonstration data.
% However, it is difficult to acquire such data for each restaurant setting.
However, collecting such data is difficult because it requires a lot of time and human resources.
Therefore, a framework for improving accuracy without using the affordance function is required.
For these reasons, we focus on understanding environments and tasks to achieve waiter tasks based on the restaurant test defined in RoboCup@Home.

\section{Proposal}
The focus of this study is waiter tasks in a restaurant, and the objective is to establish a method for facilitating HRI in real-world environments.
% Waiter tasks constitute multiple tasks, such as taking orders, serving food, and cleaning up.
Waiter tasks constitute multiple tasks, such as taking orders, serving food, and cleaning up the tables.
To ensure a robot can perform these tasks in real-world HRI, it must understand the required tasks through customer interaction.
Therefore, we propose an indoor dynamic map that separately represents static and dynamic information for HRI in dynamic environments.
Additionally, we propose a task representation-based task understanding and response generation for effective communicate with customers.

\subsection{Indoor dynamic map}
\begin{figure}[t]
    \centering
    \includegraphics[width=0.9\linewidth]{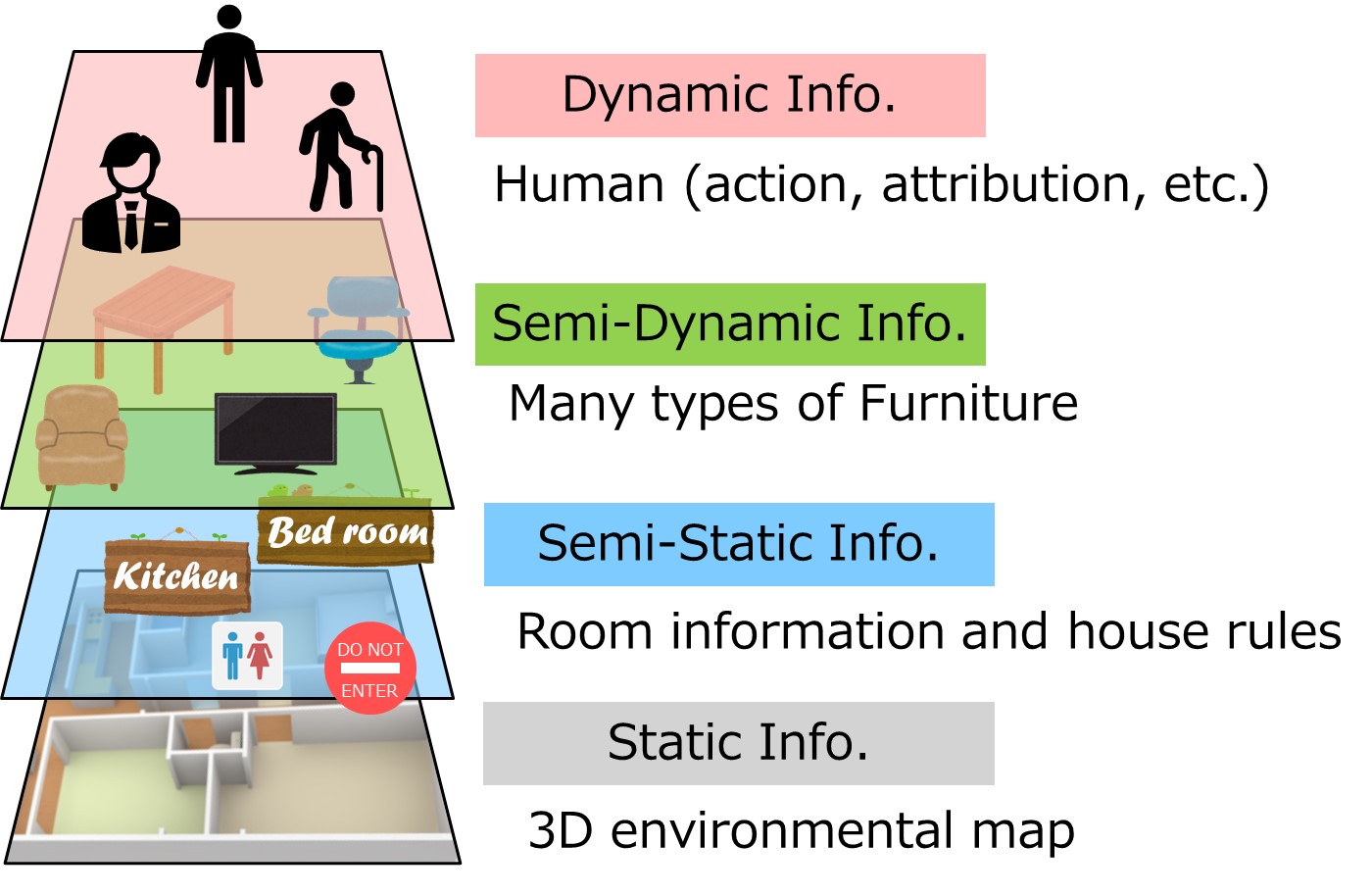}
    \caption{Proposed indoor dynamic map for embracing HRI.}
    \label{intro_dynamic_mapping}
\end{figure}

Understanding environments is necessary to optimize movements and determine destinations for service robots.
Existing environmental maps are inadequate in their expressive capabilities, making it difficult to realize the targeted waiter tasks.
In this study, we propose an indoor dynamic map for optimizing operations in real-world environments.
Fig.~\ref{intro_dynamic_mapping} shows the proposed dynamic map, which includes four layers of information based on update rates: static, semi-static, semi-dynamic, and dynamic information.

\subsubsection{Static and semi-static information}
\begin{figure}[t]
    \centering
    \includegraphics[width=\linewidth]{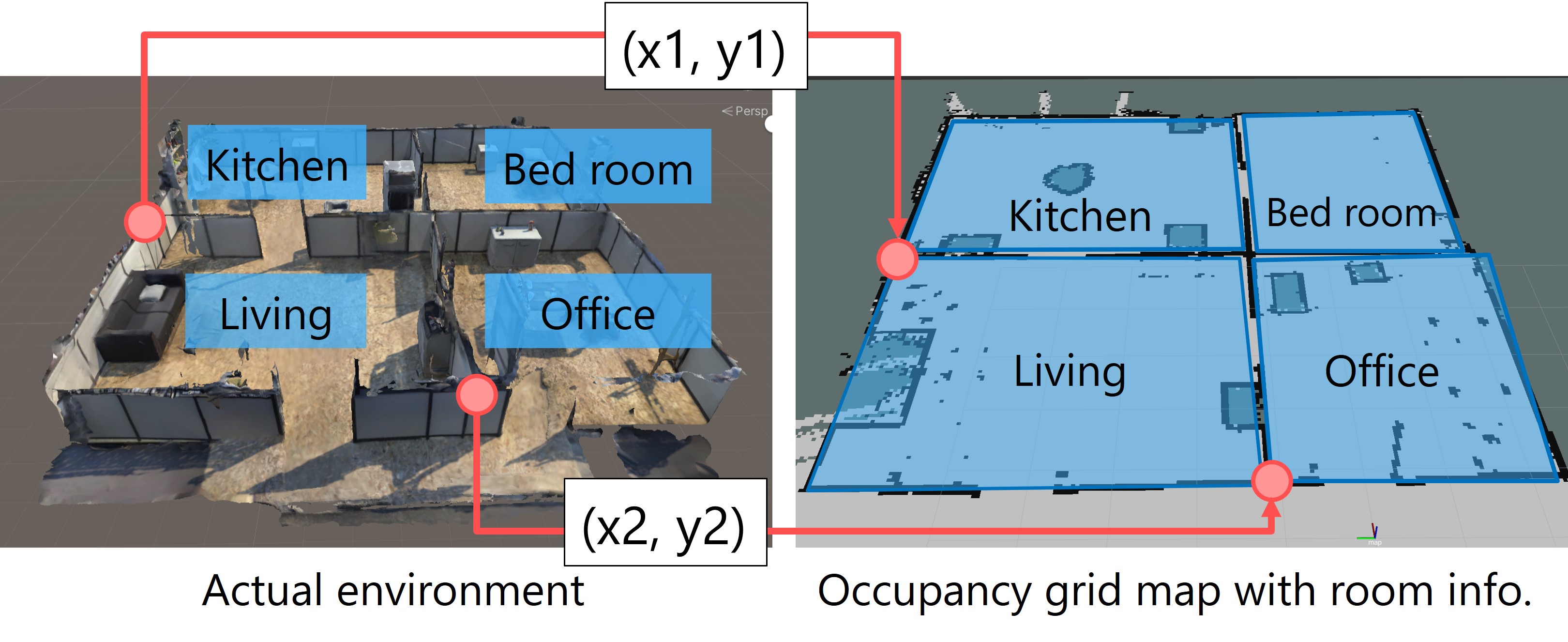}
    \caption{Static and semi-static information: Room information is incorporated through manual mapping to the occupancy grid map}
    \label{static_information}
\end{figure}

Fig.~\ref{static_information} shows the proposed static and semi-static information.
The semantic meaning of the area is specified by identifying two diagonal points on the occupancy grid map.

\subsubsection{Semi-Dynamic Information}
\begin{figure}[t]
    \centering
    \includegraphics[width=0.89\linewidth]{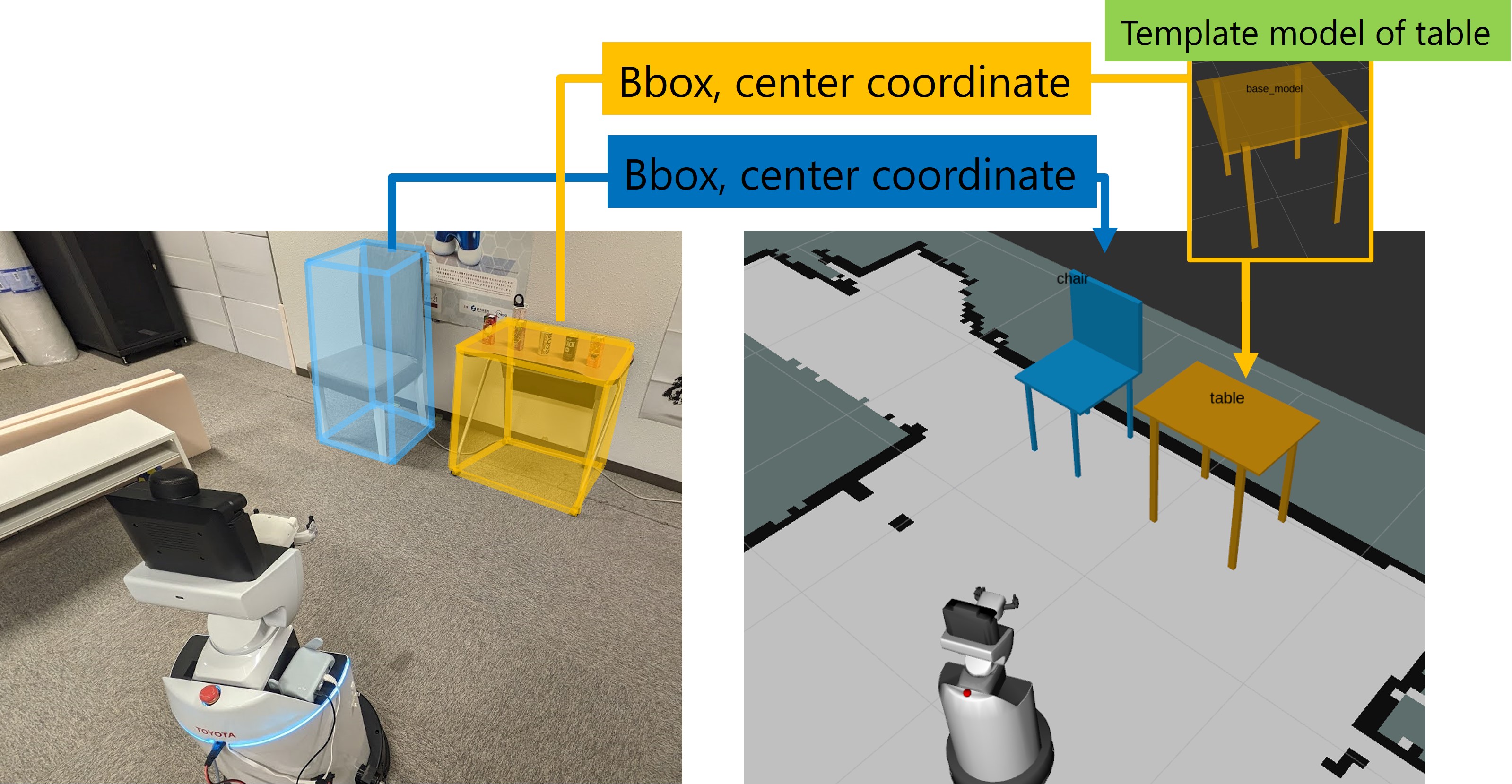}
    \caption{Semi-dynamic information: proposed furniture mapping system based on the template model.}
    \label{omni3d_recog_result}
\end{figure}

Furniture in a room may change position or its form through replacement.
Therefore, it is desirable to manage them separately from occupancy grid maps.
In this study, we define furniture as semi-dynamic information and propose a mapping method to occupancy grid maps using a furniture recognition model.
Fig.~\ref{omni3d_recog_result} shows the recognition and mapping results.
{Omni3D}~\cite{brazil2023omni3d}, the recognition method used in this study, obtains the center coordinates and measurements of 3D objects from an RGB image.
However, the rectangular bounding boxes generated by this approach cannot represent the shapes of desks, chairs, and other objects.
Therefore, we prepare a template model sized $1m \times 1m \times 1m$ for each type of furniture and load it with scale information to map furniture more clearly onto an environmental map.

Additionally, we develop a furniture registration and tracking function to reference previously recognized furniture.
We track furniture by detecting overlaps between 3D bounding boxes in the current and previous frames.
Any IDs can be assigned to the registered furniture, and information regarding the target furniture can be obtained by specifying these IDs.
If the ID is not specified during registration, the ID is automatically assigned, such as table\_0 and table\_1.

\begin{figure}[t]
    \centering
    \includegraphics[width=0.88\linewidth]{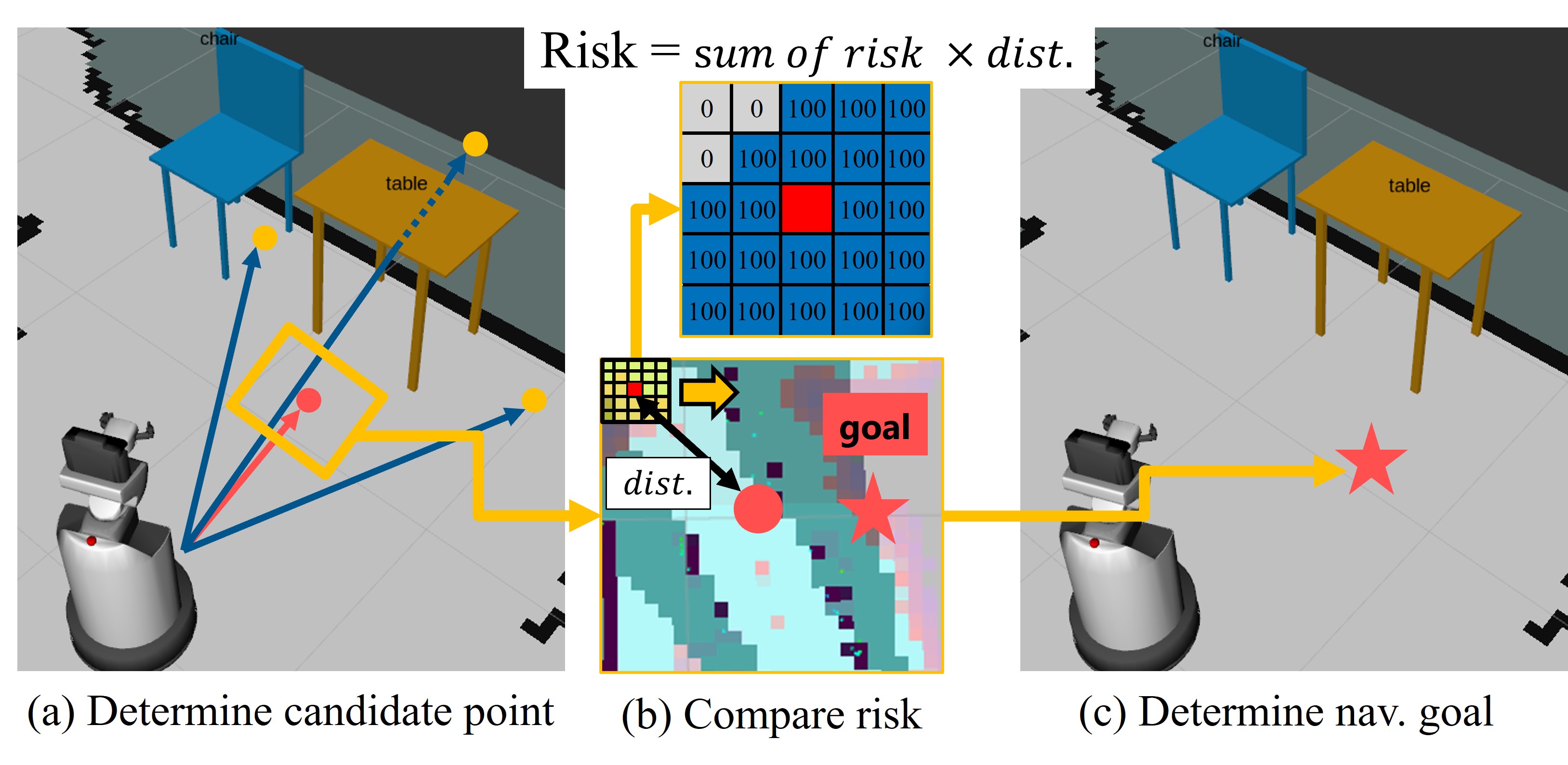}
    \caption{Determination of navigation goal based on registered furniture}
    \label{calc_nav_goal}
\end{figure}
A robot must determine its navigation goal in a dynamic environment where furniture positions change.
The proposed map determines a navigation goal based on information regarding furniture registered on the proposed map.
Fig.~\ref{calc_nav_goal} shows an overview of the proposed process for calculating navigation goals.
First, a robot refers to the four points surrounding the registered furniture and considers the point closest to its own position as a candidate point, as shown in Fig.~\ref{calc_nav_goal}(a).
Next, we compare risk values associated with each grid to determine the navigation goal.
In this study, we set the risk value of the grid within the radius of the robot bogie from the obstacle as 100.
% In this study, we set the risk value for all grids within a radius r of the obstacle to 100. 
In addition, the risk value is weighted according to the distance to select a closer grid, as shown in Fig.~\ref{calc_nav_goal}(b).
Finally, a robot calculates the sum of the risk values of the surrounding grids for each grid.
A robot selects the grid with the lowest cost as the navigation goal, as shown in Fig.~\ref{calc_nav_goal}(c).
% The proposed method delivers virtual obstacles to the grid where furniture is registered, enabling safer navigation.
The proposed method publishes virtual obstacles to the grid where furniture is registered, enabling safer navigation.

The registered furniture is also registered in the collision world of HSR, enabling the calculation of arm paths that avoid obstacles.
Because the indoor dynamic map retains the detailed shape of the furniture, it can also realize movements, such as grabbing an object under a desk.

\subsubsection{Dynamic information}
\begin{figure}[t]
    \centering
    \includegraphics[width=0.95\linewidth]{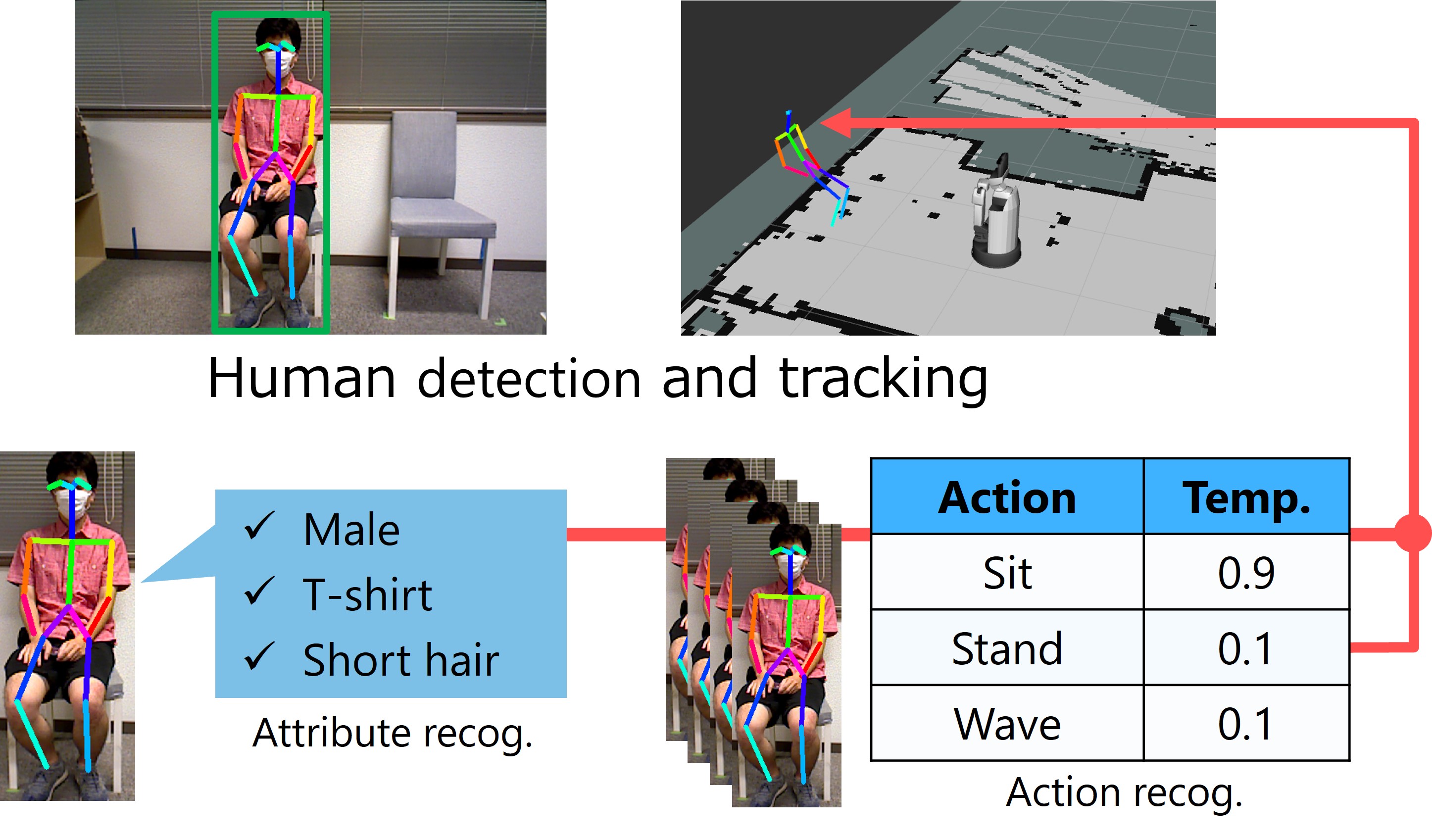}
    \caption{Dynamic information in the proposed indoor dynamic map}
    \label{dynamic_info}
\end{figure}
% Collecting information regarding human positions and attributes is essential to accomplishing flexible HRI tasks.
Collecting human positions and attributes is essential to accomplishing flexible HRI tasks.
In this study, we propose a method for mapping people's position and attributes that integrates human recognition, action recognition, and attribute estimation.
We use MMDetection \cite{mmdetection} and MMPose \cite{mmpose2020} to map a person's skeletal information onto a map, enabling us to estimate the actions of the recognized person, such as sitting, and waving.
In addition, we estimate human attributes, such as gender and clothing.
Fig.~\ref{dynamic_info} shows the results of estimating a person's 3D position, attributes, and actions.
Combining information from other layers enables the expression of a person as ``Mr. Smith is sitting on the chair in the living room, wearing a T-shirt and waving his hand.''

% \subsection{Parsing task and response generation}
\subsection{Task understanding and response generation}
\begin{figure}[t]
    \centering
    \includegraphics[width=0.95\linewidth]{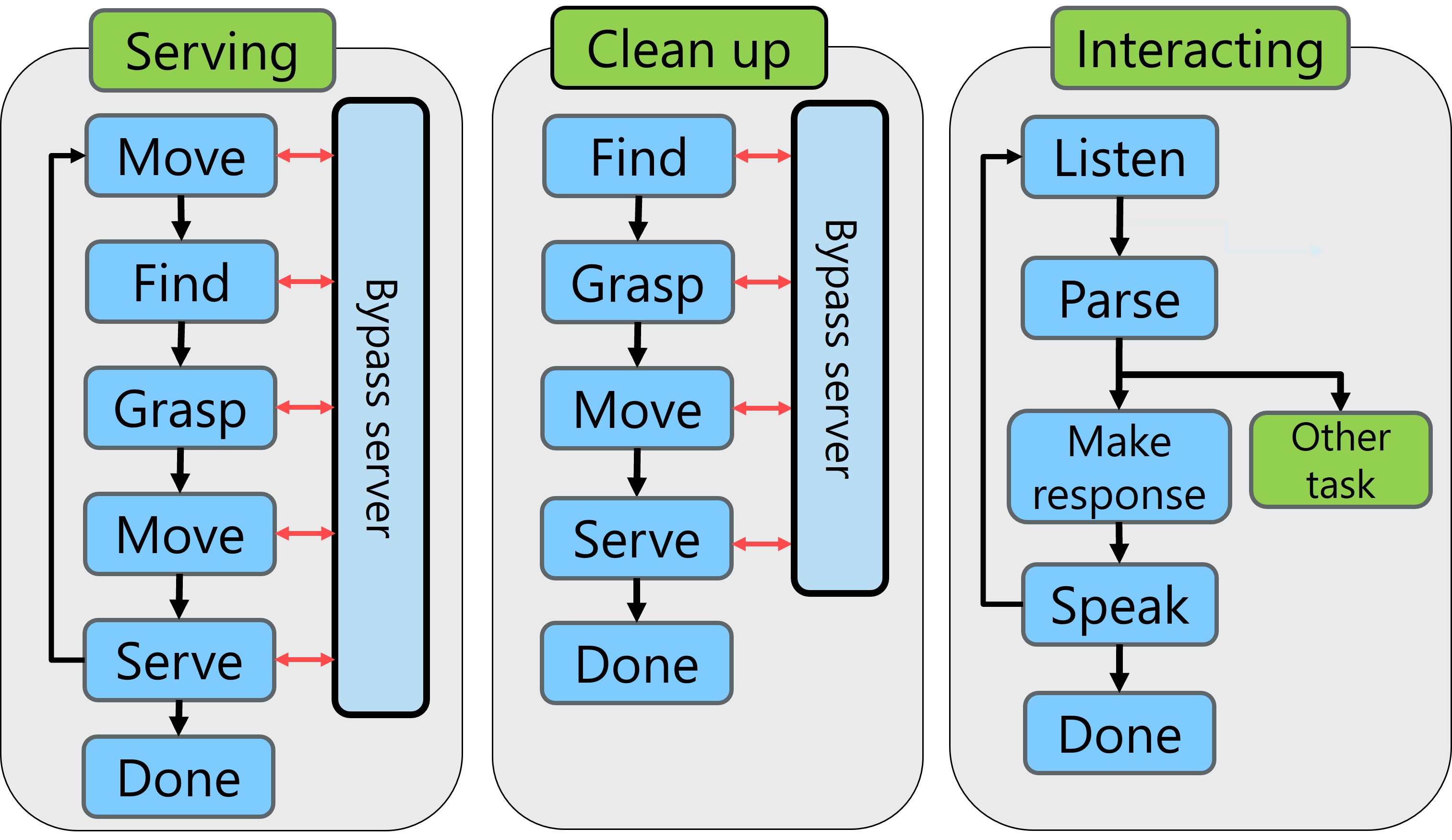}
    \caption{List of defined task representations and corresponding sequences of actions}
    \label{task_representation_details}
\end{figure}

\begin{figure}[t]
    \centering
    \includegraphics[width=0.95\linewidth]{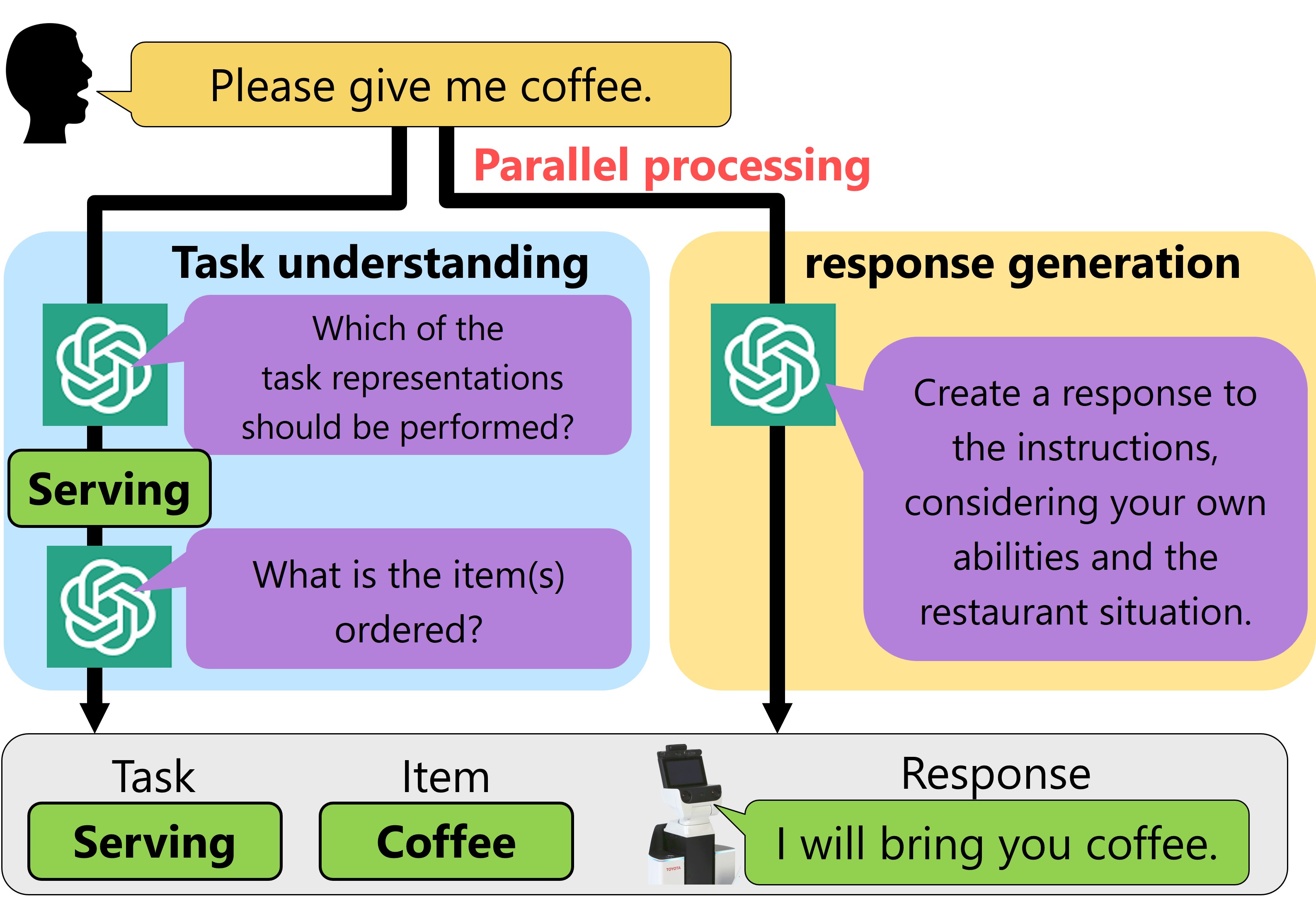}
    \caption{Overview of proposed task understanding and response generation system}
    \label{parser_system}
\end{figure}

We propose a method for understanding complex tasks in limited settings, such as restaurants.
SayCan has some limitations, such as the requirement for a large scale of data and detection accuracy decreases as the task becomes more complex.
% To address these issues, we propose a task representation method that summarizes sequences of actions to improve the accuracy of complex tasks.
To address these issues, we propose a task representation method that summarizes sequences of actions to improve the accuracy of complex tasks.
Fig. \ref{task_representation_details} and \ref{parser_system} show an overview of our task understanding and response generation system.
% We define a list of task representations and a sequence of low-level actions required to realize them.
We define a list of task representations and a sequence of actions required to realize them.
We provide the robot's operating environment and a list of task representations as base prompts for GPT-4~\cite{gpt4}.
GPT-4 predicts the task representation to be performed from the base prompts and user instructions.
The robot completes the task by sequentially executing the actions defined in the predicted task representations.

However, the proposed method cannot complete a task when robots fail to execute a skill due to unexpected issues.
To address this problem, we propose a bypass server that generates speech messages asking humans for help based on failure details.
The bypass server is implemented using GPT-4, which provides a prompt with failure details and examples of appropriate responses.

In addition, we propose a method for generating response to instructions using GPT-4.
The proposed method processes task understanding and response generation in parallel to accelerate the process.
The base prompts used for task understanding and response generation are common, and we added prompts that change only the final output format.

\section{Experiments}

\subsection{Experimental settings}
\begin{figure}[t]
    \centering
    \includegraphics[width=0.95\linewidth]{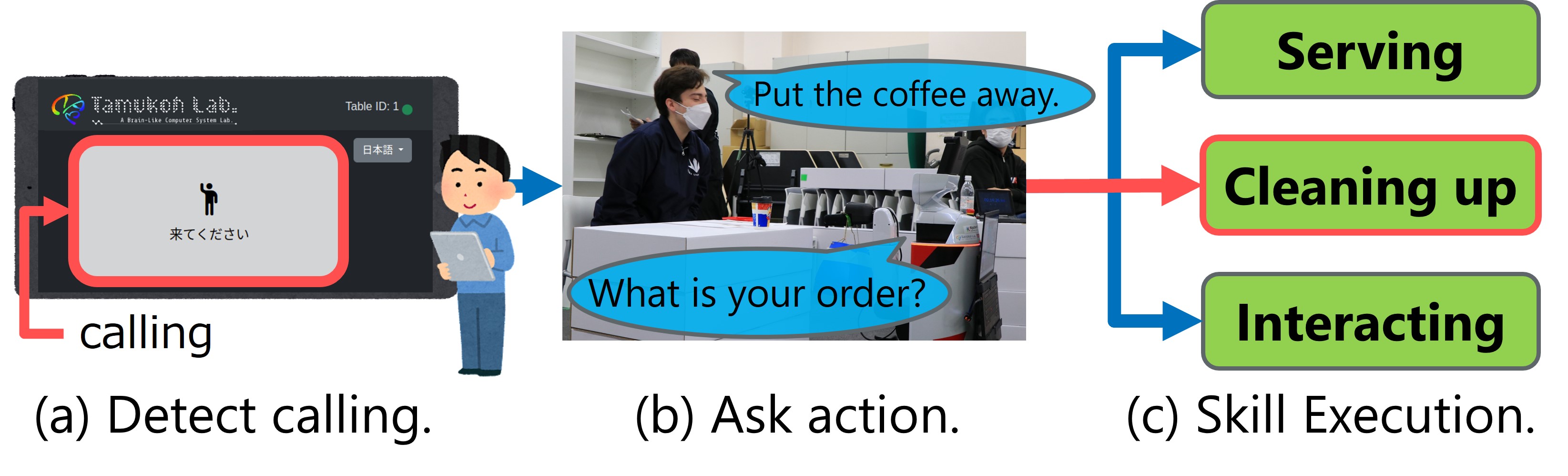}
    \caption{Designed waiter task flow}
    \label{task_flow}
\end{figure}
To evaluate the effectiveness of the proposed system, we implemented it on HSR to autonomously perform waiter tasks in an environment simulating a restaurant.
Fig. \ref{task_flow} shows flows of pre-defined waiter tasks, including
understanding orders, serving food, cleaning tables, and interacting with customers.
These task designs were based on the restaurant test in RoboCup@Home and the results of a preliminary experiment in which the robot only served food \cite{Waitersystem2023}.
Specifically, the robot responded to calls from the tables and navigated to the appropriate table.
As shown in Fig. \ref{task_flow}(a), we developed a web interface to allow customers to call a robot and distributed a tablet to each table for this purpose.
After the robot approached the table, the robot interacted with customers to understand the required task.
Finally, the robot executed multiple skills defined in the corresponding task representation.

\begin{figure}[t]
    \centering
    \includegraphics[width=0.85\linewidth]{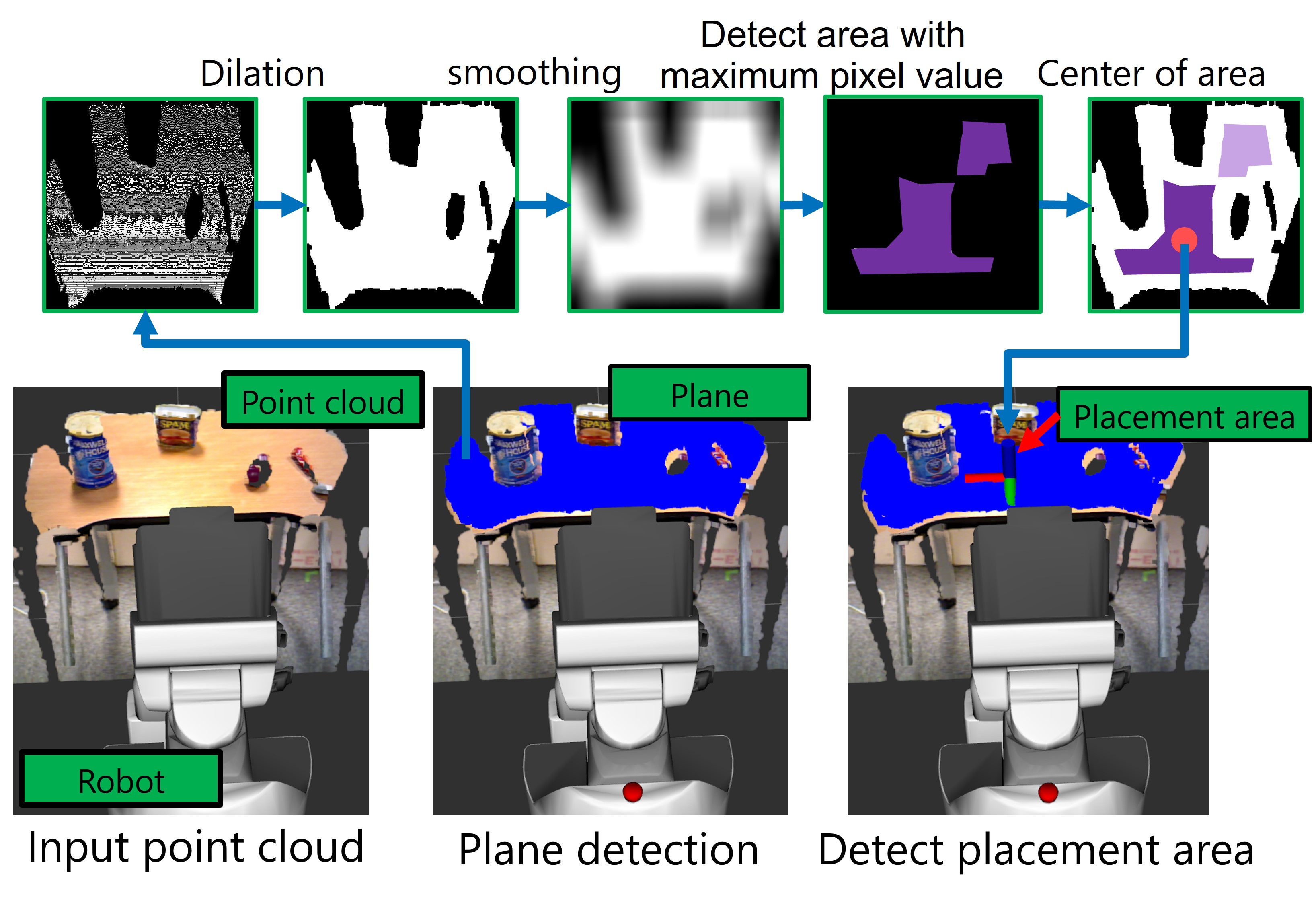}
    \caption{Placement area detection based on RANSAC}
    \label{placement_area_detection}
\end{figure}

To achieve these skills, we implemented functionalities for object grasping, placement area estimation, and speech recognition.
We used language-segment-anything (Lang-SAM) \cite{lang_sam} for object recognition, a method that can handle language information as prompts.
Lang-SAM enables the detection of ordered items without learning to adapt to the environment.
Additionally, we developed a placement area detection method based on RANSAC~\cite{ransac}, as shown in Fig. \ref{placement_area_detection}.
We implemented a real-time speech recognition system combining Silero-VAD~\cite{Silero-vad} and Whisper~\cite{Whisper}.

\begin{figure*}[t]
    \centering
    \includegraphics[width=0.90\linewidth]{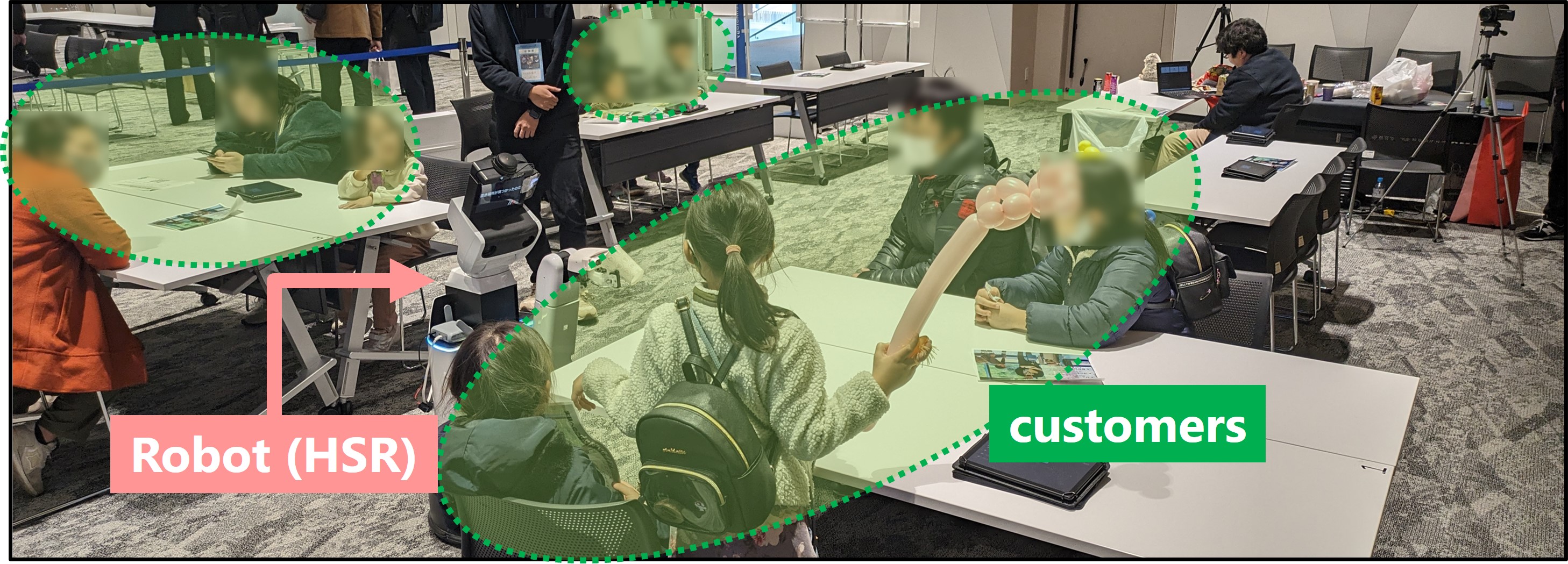}
    \caption{Overview of experimental fields}
    \label{experimental_setting}
\end{figure*}

We conducted the experiment in an environment simulating a restaurant.
Fig.~\ref{experimental_setting} shows the experimental fields.
In this experiment, we used a booth at an event to verify whether the robot could understand the environment and tasks and perform them in an environment that was not tailored for robots.
Approximately 100 people, including children, participated in the experiment.
We conducted a questionnaire survey to evaluate whether HSR equipped with our HRI system appropriately performed as a waiter.
Because people of all ages participated in the experiment, we evaluated the social acceptance of the proposed HRI system separately for each age group, expecting the social acceptance of the service robots to differ by age.

\subsection{Experimental results}
\begin{figure}[t]
    \centering
    \includegraphics[width=0.83\linewidth]{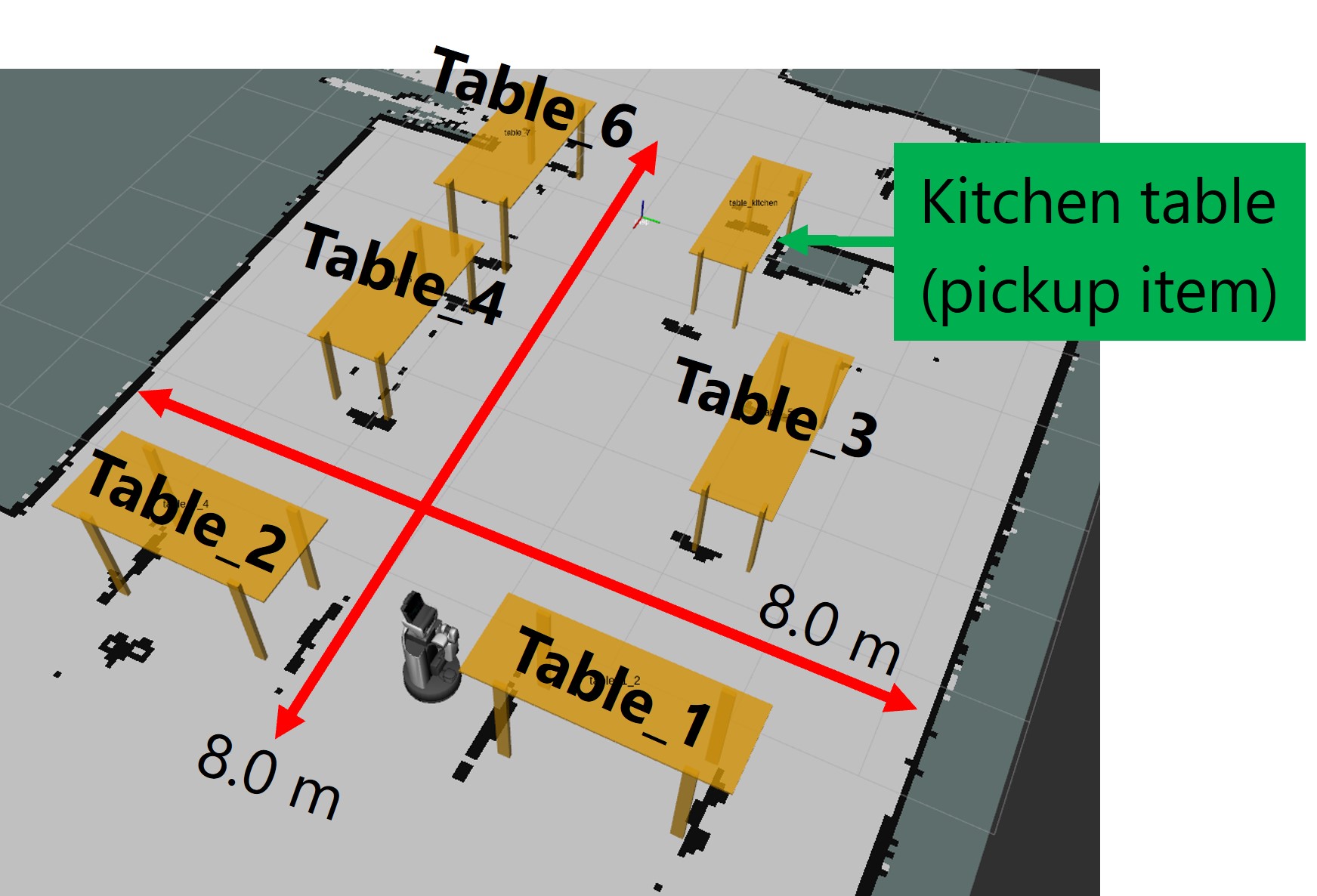}
    \caption{Created dynamic map: we only map these tables}
    \label{created_dynamic_map}
\end{figure}
We conducted the experiment for about four hours, and HSR performed 41 order-taking and serving tasks.
Out of the 41 times of serving orders, HSR delivered the incorrect item on four times.
% The remaining 37 times HSR successfully served the requested items, with some help from others.
The remaining 37 times HSR successfully served the requested items, with some help from operators and customers.
The main reason for requesting assistance was the failure to detect the desired item, prompting the robot to ask for the item to be placed in its hand to proceed with the task.
Although the customers did not ask the robot to clean the tables, some customers enjoyed in casual conversation, such as asking for a description of the items in the restaurant.
Fig.~\ref{created_dynamic_map} shows semi-dynamic information of the map created in this experiment.
The proposed method correctly detected six tables in the experimental field. In this experiment, we defined one table as a kitchen table for picking up items.
HSR navigated each table appropriately using the created indoor dynamic map.
Moreover, HSR completed the entire experiment without colliding with any obstacle.
These results indicated that the proposed system could understand commands and perform tasks with an accuracy of over 90\%.

\subsection{Questionnaire}
\begin{figure}[t]
    \centering
    \includegraphics[width=0.84\linewidth]{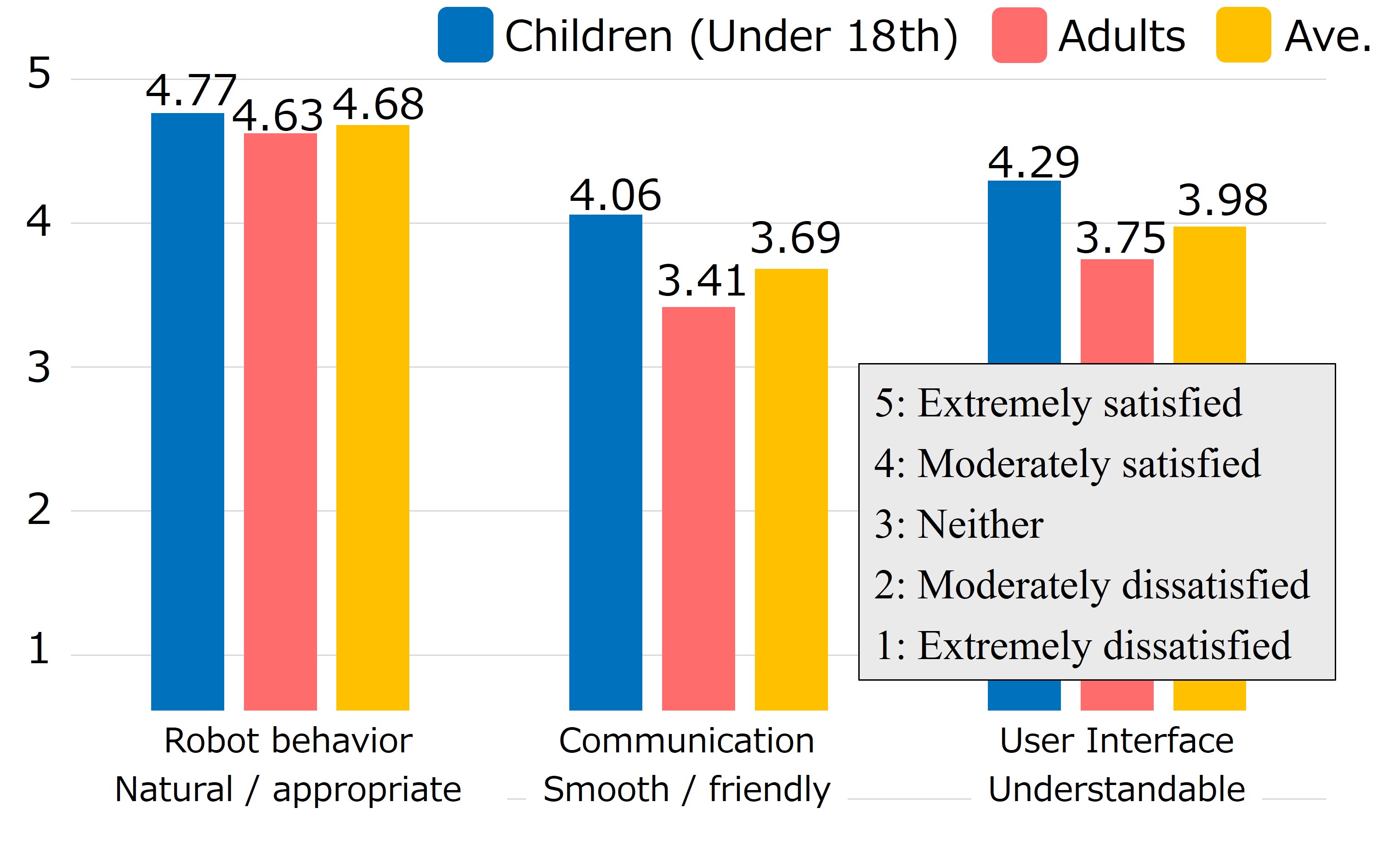}
    \caption{Results of the questionnaire survey}
    \label{questionnaire_results}
\end{figure}

We conducted the questionnaire survey on the robot performance during the experiment.
The total number of responses to the questionnaire was 41.
Fig.~\ref{questionnaire_results} shows the questionnaire results by age group.
Seventeen respondents were high-school students or younger, and 24 were university students or older.
The robot behavior received high ratings from high-school students and younger individuals, averaging 4.77 points, as well as from college students and older, averaging 4.63 points, with an overall average rating of 4.68 points.
These results indicated that the robot could identify a safe route by setting up a collision world and distributing virtual obstacles during navigation.
The robot speech content received 4.06 points from high-school students and younger individuals and 3.41 points from university students and older individuals, with an overall average of 3.69 points.
The scores did not improve likely because of instances in which the robot's speech did not match its actions.
The proposed method used separate servers for task understanding and response generation, which may lead to different interpretations in each case.
% The robot UI scores were 4.29 points for high school students and younger, 3.75 points for university students and older, and 3.98 points on average.
% Overall, the scores given by college students and above tended to be lower than those given by high school students and below.

Additionally, we conducted a questionnaire survey regarding the speed of movement and conversation.
% The average score was 2.83 points, which was below the overall average.
The average score of the movement speed was 2.83 points.
This lower score may be due to the long time required to search for grasping objects and estimate possible placement locations.
The speed of the conversation was evaluated to be appropriate, scoring 2.93 points on a scale of 1 (too slow) to 5 (too fast).
These results indicates that parallelization of response generation effectively facilitated smooth communication.
\section{Discussion}

\subsection{Indoor dynamic map}
In the experiments, the proposed method successfully generated the indoor dynamic map capable of correctly recognizing unknown tables.
HSR could move to the appropriate table and avoid collisions with obstacles based on the registered furniture.
These results indicate that the proposed indoor dynamic map can effectively realize HRI in dynamic environments.
Experimental results also show that the proposed system can work in environments with unknown objects.
However, if the shape of a desk or other furniture differs significantly from the base model, our system cannot accurately represent it on the map.
In future works, we will utilize a furniture skeleton estimation method~\cite{pose_anything}.

\subsection{Failure detection from customer feedback}
% In the experiment, even when HSR detected not-ordered foods, HSR served them to the customer.
In the experiment, even when HSR detected not-ordered foods as ordered foods, HSR served them to the customers.
One of its factors is that our system doesn't detect whether the detected objects are desirable.
In future works, we will employ a miss-recognition detector, which detects whether target objects are detected or not.
As a simple method, we consider the use of customer feedback.
% In the experiment, HSR served incorrect foods four times.
% Our system doesn't have a miss-recognition detector.
% Then, even if not-targeted objects are detected, HSR serves them to the customer.
% In future works, we will employ a miss-recognition detector, which detects whether target objects are detected or not.
% As a simple method, we consider the use of customer feedback.

% \subsection{Bypass Server}
% In the experiment, HSR served incorrect foods four times.
% The bypass server handles cases where HSR fails to reach the navigation goal or find the target object.
% However, the bypass server cannot cope with the misrecognition of speech content or objects.
% Therefore, our system must check whether the robot has completed its task from any other information.
% In future works, we will discuss a novel method in which robots collect customer feedback automatically through their facial expression and speech contents after the robot's action. 

% On the other hand, the bypass server cannot adapt with cases where the robot mistakenly believes that it has grasped a suitable product or that it has successfully navigated due to a gap in self-position estimation.
% We will discuss a method that is needed to verify whether or not the robot has really succeeded by referring to the external information after each skill execution.

\subsection{Understanding task and generating response}
In the experiment, HSR understood the task with an accuracy of over 90\% and delivered the correct item to customers.
In response to questions such as what items are available in the restaurant, the response generation system created sentences from the product information in the base prompts and introduced them to the customers.
These results indicate that the proposed task understanding and response generation system is adequate for real-world HRI.
% In addition, our system can add other tasks, such as guiding the user to an empty seat, by defining the sequence of skills such as move and find.
On the other hand, some participants indicated that the robot speech was unnatural in several situations.
One reason for this issue was that there were situations in which the robot speech and actual actions were out of sync, such as when it served items other than the ones that had been ordered or when it started serving an order while listening to the order again.

The reason for the mismatch was that the robot executed its understanding of the product after receiving the order and generating the utterance to repeat the order in parallel.
% The reason for the mismatch between the robot's understanding of the order and the content of the utterance was that the robot executed its understanding of the product after receiving the order and generating the utterance to repeat the order in parallel.
To address this problem, the system can be modified to generate utterances based on the results obtained from understanding the order.
However, the proposed method uses GPT for both order understanding and generation of response, which is undesirable because significant time is required to generate a response.
In future works, we will discuss replacing order understanding with a system based on a small-scale language model, such as GPT4ALL \cite{gpt4all}, to create a faster and more accurate response system. 

\subsection{Questionnaire}
HSR, equipped with the proposed HRI system, received favorable ratings (4.68 points) for its performance as a waiter.
These results indicates that the task understanding and response generation systems can effectively accomplish real-world HRI.
In this experiment, the web interface was prepared for participants to call the robot to the table.
Six out of 41 participants indicated that they wanted to complete the ordering process in addition to calling the robot. 
Only one respondent was a high school student or younger.
These results indicate that voice interaction is essential for children, and
the proposed system can allow children to operate the robot intuitively by voice.

\section{Conclusion}
In this study, we proposed the indoor dynamic map, the task representation-based task understanding system, and the response generation system for HRI.
In addition, we developed the flexible HRI system with proposed methods.
To verify the effectiveness of this system, we conducted the experiment with approximately 100 people, including children, in an environment simulating a restaurant.
The experimental results show that the proposed system successfully understand commands and serve food with more than 90\% accuracy.
In addition, the questionnaire results shows that the robot behavior and HRI received more than four points, indicating favorable reception by participants of all ages.
In future works, we will continue to improve each proposal to adapt it to more complex home environments.

% 本研究では，屋内での動作に向けたダイナミックマップの提案と，マップを活用したHRIシステムの開発を行った．
% 有効性検証のために，我々はレストランを模した環境にて子供を含む100人の方を対象として実験を行った．
% 実験結果より，90\%以上の精度でのコマンド理解と配膳に成功した．
% また，アンケート結果より，ロボットの動作やHRIに関する評価が4ポイント以上であり，広い年代の人に対して，提案システムが好意的にとらえられた．
% 今後は，さらに複雑な家庭環境への適応を目指した各提案の改良を進める．

\addtolength{\textheight}{-12cm}

% \section*{ACKNOWLEDGMENT}
% This paper is based on results obtained from a project, JPNP16007, commissioned by the New Energy and Industrial Technology Development Organization (NEDO).
% This work was supported by JSPS KAKENHI Grant Numbers 23H03468, 23K18495.

\bibliographystyle{ieeetr}
\bibliography{main}

\begin{thebibliography}{10}

\bibitem{Waitersystem2023}
Y.~Yano, K.~Isomoto, T.~Ono, and H.~Tamukoh, ``{Autonomous Waiter Robot System for Recognizing Customers, Taking Orders, and Serving Food},'' in {\em RoboCup 2023: Robot World Cup XXVI}, pp.~252--261, 2024.

\bibitem{Ono2022}
T.~Ono, D.~Kanaoka, T.~Shiba, S.~Tokuno, Y.~Yano, A.~Mizutani, I.~Matsumoto, H.~Amano, and H.~Tamukoh, ``{Solution of World Robot Challenge 2020 Partner Robot Challenge (Real Space)},'' vol.~36, no.~17, pp.~870--889, 2022.

\bibitem{realworld_robotics}
{Center for Research and Development Strategy}, ``Real-world robotics a new principle for robotics that flexibly adapts to open environments.'' \url{https://www.jst.go.jp/crds/pdf/2022/SP/CRDS-FY2022-SP-02.pdf}.
\newblock Accessed Jan. 3 2024.

\bibitem{Rosinol19icra-incremental}
A.~Rosinol, T.~Sattler, M.~Pollefeys, and L.~Carlone, ``Incremental visual-inertial 3d mesh generation with structural regularities,'' in {\em Proceedings of the 2019 International Conference on Robotics and Automation (ICRA)}, pp.~8220--8226, 2019.

\bibitem{Rosinol20icra-Kimera}
A.~Rosinol, M.~Abate, Y.~Chang, and L.~Carlone, ``{Kimera}: an open-source library for real-time metric-semantic localization and mapping,'' in {\em Proceedings of the 2020 International Conference on Robotics and Automation (ICRA)}, pp.~1689--1696, 2020.

\bibitem{saycan2022arxiv}
M.~Ahn, A.~Brohan, N.~Brown, Y.~Chebotar, O.~Cortes, B.~David, C.~Finn, C.~Fu, K.~Gopalakrishnan, K.~Hausman, A.~Herzog, D.~Ho, J.~Hsu, J.~Ibarz, B.~Ichter, A.~Irpan, E.~Jang, R.~J. Ruano, K.~Jeffrey, S.~Jesmonth, N.~J. Joshi, R.~Julian, D.~Kalashnikov, Y.~Kuang, K.-H. Lee, S.~Levine, Y.~Lu, L.~Luu, C.~Parada, P.~Pastor, J.~Quiambao, K.~Rao, J.~Rettinghouse, D.~Reyes, P.~Sermanet, N.~Sievers, C.~Tan, A.~Toshev, V.~Vanhoucke, F.~Xia, T.~Xiao, P.~Xu, S.~Xu, M.~Yan, and A.~Zeng, ``{Do As I Can, Not As I Say}: Grounding language in robotic affordances,'' {\em arXiv:2204.01691}, 2022.

\bibitem{mulit_layer}
P.-T. Wu, C.-A. Yu, S.-H. Chan, M.-L. Chiang, and L.-C. Fu, ``Multi-layer environmental affordance map for robust indoor localization, event detection and social friendly navigation,'' in {\em Proceedings of the 2019 {IEEE/RSJ} International Conference on Intelligent Robots and Systems (IROS)}, pp.~2945--2950, 2019.

\bibitem{yolov3}
J.~Redmon and A.~Farhadi, ``{YOLOv3}: An incremental improvement,'' {\em arXiv:1804.02767}, 2018.

\bibitem{robocup}
``{RoboCup@Home}.''
\newblock \url{http://www.robocupathome.org} (Accessed 18. Jan. 2024).

\bibitem{yamamoto2019hsr}
T.~Yamamoto, K.~Terada, A.~Ochiai, F.~Saito, Y.~Asahara, and K.~Murase, ``Development of human support robot as the research platform of a domestic mobile manipulator,'' {\em ROBOMECH journal}, vol.~6, no.~4, 2019.

\bibitem{brazil2023omni3d}
G.~Brazil, A.~Kumar, J.~Straub, N.~Ravi, J.~Johnson, and G.~Gkioxari, ``{Omni3d}: A large benchmark and model for 3d object detection in the wild,'' in {\em Proceedings of the 2023 {IEEE/CVF} conference on computer vision and pattern recognition (CVPR)}, pp.~13154--13164, 2023.

\bibitem{mmdetection}
K.~Chen, J.~Wang, J.~Pang, Y.~Cao, Y.~Xiong, X.~Li, S.~Sun, W.~Feng, Z.~Liu, J.~Xu, Z.~Zhang, D.~Cheng, C.~Zhu, T.~Cheng, Q.~Zhao, B.~Li, X.~Lu, R.~Zhu, Y.~Wu, J.~Dai, J.~Wang, J.~Shi, W.~Ouyang, C.~C. Loy, and D.~Lin, ``Mmdetection: Open mmlab detection toolbox and benchmark,'' {\em arXiv:1906.07155}, 2019.

\bibitem{mmpose2020}
{MMPose Contributors}, ``Openmmlab pose estimation toolbox and benchmark.'' \url{https://github.com/open-mmlab/mmpose}, 2020.
\newblock Accessed 05 Jan. 2024.

\bibitem{gpt4}
``{OpenAI GPT-4}.'' \url{https://openai.com/research/gpt-4}.
\newblock Accessed 5. Feb. 2024.

\bibitem{lang_sam}
``lang-segment-anything.'' \url{https://github.com/luca-medeiros/lang-segment-anything}.
\newblock Accessed 5. Feb. 2024.

\bibitem{ransac}
R.~Raguram, O.~Chum, M.~Pollefeys, J.~Matas, and J.-M. Frahm, ``{USAC}: A universal framework for random sample consensus,'' in {\em IEEE Transactions on Pattern Analysis and Machine Intelligence (PAMI)}, vol.~35, pp.~2022--2038, 2013.

\bibitem{Silero-vad}
{Silero Team}, ``Silero {VAD}: pre-trained enterprise-grade voice activity detector ({VAD}), number detector and language classifier.'' \url{https://github.com/snakers4/silero-vad}.
\newblock Accessed 5. Feb. 2024.

\bibitem{Whisper}
A.~Radford, J.~W. Kim, T.~Xu, G.~Brockman, C.~McLeavey, and I.~Sutskever, ``Robust speech recognition via large-scale weak supervision,'' in {\em Proceedings of the 40th International Conference on Machine Learning (ICML)}, pp.~28492--28518, 2023.

\bibitem{pose_anything}
O.~Hirschorn and S.~Avidan, ``{Pose Anything: A Graph-Based Approach for Category-Agnostic Pose Estimation},'' {\em arxiv:2311.17891}.

\bibitem{gpt4all}
Y.~Anand, Z.~Nussbaum, B.~Duderstadt, B.~Schmidt, and A.~Mulyar, ``{GPT4All}: Training an assistant-style chatbot with large scale data distillation from {GPT-3.5-Turbo}.'' \url{https://github.com/nomic-ai/gpt4all}.
\newblock Accessed 05 Feb. 2024.

\end{thebibliography}

\end{document}